# Solutions for Practice-oriented Requirements for Optimal Path Planning for the AUV "SLOCUM Glider"

M. Eichhorn Member IEEE
Institute for Ocean Technology
National Research Council Canada
Arctic Avenue, P.O. Box 12093
St. John's, Newfoundland A1B 3T5, Canada
phone +01 709-772-7986; fax + 01 709-772-2462
e-mail: mike.eichhorn@nrc-cnrc.gc.ca

*Abstract*- **This paper presents a few important practice-oriented requirements for optimal path planning for the AUV "SLOCUM Glider" as well as solutions using fast graph based algorithms. These algorithms build upon the TVE (time-varying environment) search algorithm. The experience with this algorithm, requirements of real missions along the Newfoundland and Labrador Shelf and the idea to find the optimal departure time are the motivation to address the field of research, which is described in this paper. The main focus of this paper is a discussion of possible methods to accelerate the path planning algorithm, without deterioration of the results.**

## I. INTRODUCTION

The solutions and algorithms presented in this article are focused on path planning requirements for the AUV "SLOCUM" glider, which is a particular type of autonomous underwater vehicle (AUV). These gliders have a low cruising speed (0.2 to 0.4 m s$^{-1}$) in a time-varying ocean flow for long operational periods up to 30 days. The algorithms presented here are equally applicable to other AUV's as well as land based or aerial mobile autonomous systems. The practice-oriented requirements discussed in this paper in detail are:

- fast calculation of the path planning algorithm;
- detection of the optimal departure time; and
- finding a path through the region of interest where the vehicle can collect as much oceanographic data as possible under consideration of the adverse ocean current.

The path planning algorithms presented in this paper build upon a graph algorithm, named the TVE algorithm, which is described in [1] and [2] and which has achieved practical use in [3]. This search algorithm is based on a modified Dijkstra Algorithm (see [4] and [5]), including the time-variant cost function in the algorithm which will be calculated during the search to determine the travel times (cost values) for the examined edges. This modification allows the determination of a time-optimal path in a time-varying environment. In [5] this principle was used to find the optimal link combination to send a message via a computer communication network with the shortest transport delay.

The requirement to accelerate the existing TVE algorithm results from trying to determine real mission plans for the AUV "SLOCUM Glider" to collect oceanographic data along the Newfoundland and Labrador Shelf [3]. The calculation of the travel time for an edge requires a large amount of local ocean current information, which is generated in the ocean current model. This model uses interpolation methods to extract the current information from netCDF data files. Because the number of edges in the geometrical graph ranges from one hundred thousand to one million for a mission of duration ten days, the sum of the cost function calculations is very time-intensive. This calculation will be augmented by using an accurate glider-model [6] in the cost function. In [7] a modified A*-Algorithm was used to reduce the computational cost of the search. This algorithm, called Constant-Time Surfacing A*(CTS-A*), is designed for the glider characteristics.

A fast working algorithm is also a precondition for the second requirement, the detection of an optimal departure time. A symbolic wavefront expansion (SWE) technique for an Unmanned Air Vehicle (UAV) in time-varying winds was introduced in [8] to find the time optimal path and additionally the optimal departure time. The TVE uses a similar principle as is used in the SWE to calculate the time-varying cost function for the several vertices. This includes the arrival time at the several vertices in the cost function calculation during the search. The TVE can be understood as a particular case of the SWE, where the specified departure time is a single value. To find the optimal departure time, the SWE and the approach described in this paper use separate solution methods. The reasons are the accurate and fast determination of the optimal departure time, as well as the inclusion of uncertainties in the path planning as a result of forecast error variance, accuracy of calculation in the cost functions and a possible use of a different vehicle speed in the real mission than planned.

## II. METHODS TO REALIZE FAST SEARCH-ALGORITHMS

The work with the TVE algorithm [3] while using computing intensive cost functions to find an optimal dive profile for the glider as well as the multiple calls of the algorithm in the case of detection of the optimal departure time (see section III) demonstrates the imperative of a fast working algorithm. In this section a few methods to accelerate the processing time of the TVE algorithm will be presented.

### A. Improvement of the TVE algorithm

In [1] and [2] we described a search method based on a classical Dijkstra algorithm [4] to find an optimal path in a time-varying environment. In comparison to the Dijkstra algorithm the TVE algorithm calculates the weight for the examined edges during the search using a function *wfunc* (see section II.b in [2] for detailed explanations and the left column of TABLE I). This function calculates the travel time to drive along the edge from a start vertex *u* to an end vertex *v* using a given start time. The start time to be used will be the current cost value *d(u)*, which describes the travel time from the source vertex *s* to the start vertex *u*. The examination of the edge *(u,v)* is only necessary for *d(u) < d(v)*. This is the case when the vertex *v* will visit the first time (*d(v)* is here ∞) or when the vertex *v* was visited before and the cost value *d(u)* is smaller than *d(v)*. This modification is highlighted in the right column of TABLE I, which shows a comparison between the TVE and the improved TVE (ITVE) algorithm. It is clear if *d(u) ≥ d(v)* then $d_v > d(v)$, independent of the calculated cost value of function *wfunc*. This simple modification leads to a considerable decrease in the number of cost function calls *wfunc* during the search and so to a faster processing time of the algorithm.

### B. A* algorithm

Another possible method to accelerate the TVE algorithm is the inclusion of an A* algorithm [9]. The A* algorithm utilizes the Dijkstra algorithm and uses a heuristic function *h(u)* to decrease the processing time of the path search. As a heuristic function, the travel time $t_{travel}$ from the current node *u* to the goal node *g* following a straight line based on [7] will be used. Here the travel time will be calculated using the maximum possible speed, as determined by the addition of the vehicle speed through the water $v_{veh\_bf}$ and the maximum ocean current velocity $v_{current\_max}$ in the operational area over the full mission time:

$$h(u) = t_{travel} = \frac{\left\| \mathbf{x}_u - \mathbf{x}_g \right\|}{v_{veh\_bf} + v_{current\_max}} . \quad (1)$$

This requirement ensures that the estimated costs to the goal point are never overestimated under all conditions, which is a precondition for the A* algorithm to work correctly. TABLE II shows a comparison between the ITVE algorithm without (left column) and with inclusion of the A* algorithm (right column). The differences between the algorithms are highlighted. The using of empty lines in the left column should serve to better clarify the comparison of the several program steps. The main differences between the two algorithms are the used values in the priority queue (ITVE: cost value *d*; A*TVE: *f* as sum of cost value *d* and heuristic *h*) and the break of the while loop in the case that the extracted node *u* is the goal node *g*.

TABLE I
PSEUDO-CODE OF THE TVE AND ITVE ALGORITHMS

| TVE(G, s, t₀) | ITVE(G, s, t₀) |
|---|---|
| for each vertex u ∈ V | for each vertex u ∈ V |
|   d[u] ← ∞ |   d[u] ← ∞ |
|   π[u] ← ∞ |   π[u] ← ∞ |
|   color[u] ← WHITE |   color[u] ← WHITE |
| color[s] ← GRAY | color[s] ← GRAY |
| d[s] ← t₀ | d[s] ← t₀ |
| INSERT(Q, s) | INSERT(Q, s) |
| while (Q≠Ø) | while (Q≠Ø) |
|   u ← EXTRACT-MIN(Q) |   u ← EXTRACT-MIN(Q) |
|   color[u] ← BLACK |   color[u] ← BLACK |
|   for each v ∈ Adj[u] |   for each v ∈ Adj[u] |
|  |     if (d[u] < d[v]) |
|     $d_v$ = wfunc(u, v, d[u]) + d[u] |     $d_v$ = wfunc(u, v, d[u]) + d[u] |
|     if ($d_v$ < d[v]) |     if ($d_v$ < d[v]) |
|       d[v] ← $d_v$ |       d[v] ← $d_v$ |
|       π[v] ← u |       π[v] ← u |
|       if (color[v] = GRAY) |       if (color[v] = GRAY) |
|         DECREASE-KEY(Q,v,$d_v$) |         DECREASE-KEY(Q,v,$d_v$) |
|       else |       else |
|         color[v] ← GRAY |         color[v] ← GRAY |
|         INSERT(Q, v) |         INSERT(Q, v) |
| return (d, π) | return (d, π) |

TABLE II
PSEUDO-CODE OF THE ITVE AND A*TVE ALGORITHMS

| ITVE(G, s, t₀) | A*TVE(G, s, g, t₀) |
|---|---|
| for each vertex u ∈ V | for each vertex u ∈ V |
|   d[u] ← ∞ |   d[u] ← ∞ |
|  |   f[u] ← ∞ |
|   π[u] ← ∞ |   π[u] ← ∞ |
|   color[u] ← WHITE |   color[u] ← WHITE |
| color[s] ← GRAY | color[s] ← GRAY |
| d[s] ← t₀ | d[s] ← t₀ |
|  | f[s] ← t₀ + h[s] |
| INSERT(Q, s) | INSERT(Q, s) |
| while (Q≠Ø) | while (Q≠Ø) |
|   u ← EXTRACT-MIN(Q,d) |   u ← EXTRACT-MIN(Q,f) |
|  |   if (u = g) |
|  |     return (d, π) |
|   color[u] ← BLACK |   color[u] ← BLACK |
|   for each v ∈ Adj[u] |   for each v ∈ Adj[u] |
|     if (d[u] < d[v]) |     if (d[u] < d[v]) |
|     $d_v$ = wfunc(u, v, d[u]) + d[u] |     $d_v$ = wfunc(u, v, d[u]) + d[u] |
|     if ($d_v$ < d[v]) |     if ($d_v$ < d[v]) |
|       d[v] ← $d_v$ |       d[v] ← $d_v$ |
|  |       f[v] ← $d_v$ + h[v] |
|       π[v] ← u |       π[v] ← u |
|       if (color[v] = GRAY) |       if (color[v] = GRAY) |
|         DECREASE-KEY(Q,v,$d_v$) |         DECREASE-KEY(Q,v,f[v]) |
|       else |       else |
|         color[v] ← GRAY |         color[v] ← GRAY |
|         INSERT(Q, v) |         INSERT(Q, v) |
| return (d, π) | return (d, π) |

## C. Optimal navigation formula from Zermelo

The use of the TVE algorithm to find a time optimal path for the AUV "SLOCUM" glider in time varying ocean flows allows a further possibility to reduce the computing time of the search. This approach uses the optimal navigation formula from Zermelo [10]:

$$\frac{d\theta}{dt} = -u_y\cos^2\theta + (u_x - v_y)\cos\theta\sin\theta + v_x\sin^2\theta \quad (2)$$

with $\theta$ as the heading and $u_x$, $u_y$, $v_x$ and $v_y$ as the partial derivatives of the ocean current components $u$ and $v$. The idea to develop this formula came to Zermelo's mind when the airship "Graf Zeppelin" circumnavigated the earth in August 1929 [11]. This formula describes the necessary condition for the control law of the heading $\theta$, to steer a vehicle in a time-optimal sense through a time-varying current field. The gradient of the resulting optimal trajectory in a fixed world coordinate system is the vehicle velocity over the ground $v_{veh\_og}$. This vector is the result of a vector addition of the current vector $v_{current}$ and the $v_{veh\_bf}$ vector with vehicle speed through the water $v_{veh\_bf}$ as norm and heading $\theta$ as direction. The direction of this vector $v_{veh\_og}$ is the course over the ground (COG) $\phi$. These relationships are illustrated in Fig. 1. The idea of how to use the optimal navigation formula in the search algorithm as well as the several necessary program steps will be described subsequently.

If we assume the search algorithm will find the time-optimal path, then the several segments (edges) of this path will match well with the optimal trajectory, which is calculated with optimal control by solving the optimal navigation formula from Zermelo (see section V.A). This assumption means that during the path search only vertices should be considered where the connections (edges) comply with the optimal navigation formula. This compliance is required where the transition that is the change of direction between two adjacent edges is matched with (2). TABLE III shows a comparison between the ITVE algorithm (left column) and extensions of the program steps to involve the optimal navigation formula (right column). The additional program steps are highlighted and the main steps are described as follows:

1. Only in the case that the examined vertex $u$ is not the start vertex $s$ ($u \neq s$), the additional program steps will be executed because the calculation of the optimal course, $\phi_{opt}$, requires a predecessor vertex.

TABLE III
PSEUDO-CODE OF THE ITVE AND ZTVE ALGORITHMS

| ITVE($G$, $s$, $t_0$) | ZTVE($G$, $s$, $t_0$, $\phi_{\Delta max}$) |
|---|---|
| for each vertex $u \in V$ | for each vertex $u \in V$ |
|   $d[u] \leftarrow \infty$ |   $d[u] \leftarrow \infty$ |
|   $\pi[u] \leftarrow \infty$ |   $\pi[u] \leftarrow \infty$ |
|   color[$u$] $\leftarrow$ WHITE |   color[$u$] $\leftarrow$ WHITE |
| color[$s$] $\leftarrow$ GRAY | color[$s$] $\leftarrow$ GRAY |
| $d[s] \leftarrow t_0$ | $d[s] \leftarrow t_0$ |
| INSERT($Q$, $s$) | INSERT($Q$, $s$) |
| while ($Q \neq \emptyset$) | while ($Q \neq \emptyset$) |
|   $u \leftarrow$ EXTRACT-MIN($Q$) |   $u \leftarrow$ EXTRACT-MIN($Q$) |
|   color[$u$] $\leftarrow$ BLACK |   color[$u$] $\leftarrow$ BLACK |
| |   if ($u \neq s$) |
| |     $\phi_{opt}$=CALC-OPTDIR($\pi[u]$,$u$,$d[\pi[u]]$,$d[u]$) |
|   for each $v \in Adj[u]$ |   for each $v \in Adj[u]$ |
|     if ($d[u] < d[v]$) |     if ($d[u] < d[v]$) |
| |       $\phi_{path}$ = CALC-PATHDIR ($u$,$v$) |
| |       if (($u$=$s$) OR ($|\phi_{opt}-\phi_{path}|<\phi_{\Delta max}$)) |
|       $d_v$ = wfunc($u$, $v$, $d[u]$) + $d[u]$ |         $d_v$ = wfunc($u$, $v$, $d[u]$) + $d[u]$ |
|       if ($d_v < d[v]$) |         if ($d_v < d[v]$) |
|         $d[v] \leftarrow d_v$ |           $d[v] \leftarrow d_v$ |
|         $\pi[v] \leftarrow u$ |           $\pi[v] \leftarrow u$ |
|         if (color[$v$] = GRAY) |           if (color[$v$] = GRAY) |
|           DECREASE-KEY($Q$,$v$,$d_v$) |             DECREASE-KEY($Q$,$v$,$d_v$) |
|         else |           else |
|           color[$v$] $\leftarrow$ GRAY |             color[$v$] $\leftarrow$ GRAY |
|           INSERT($Q$, $v$) |             INSERT($Q$, $v$) |
| return ($d$, $\pi$) | return ($d$, $\pi$) |

2. If the examined vertex $u$ is the start vertex $s$ all successor vertices, or rather all *edges (u,v)*, will be examined. Thereby as well as through the defined angle range $\pm\Delta\phi_{max}$ (see point 4) we develop a branching structure of possible paths during the search which is comparable with a tree with $s$ as root.

3. The direction of the edge (path element) with the predecessor vertex $\pi[u]$ as start vertex and the current examined vertex $u$ as end vertex should reflect the average optimal course over ground $\phi$ of the vehicle if it drives along this edge. This direction will be used in the function CAL-OPTDIR (for details see section D) to calculate the optimal path direction $\phi_{opt}$ for the subsequent edge on position $u$ to the time $d(u)$ based of the optimal trajectory calculated with the optimal navigation formula from Zermelo.

4. The calculated path direction $\phi_{opt}$ will be used to select possible successor edges with the end vertex $v$ under consideration of an angle range $\pm\Delta\phi_{max}$ (see Fig. 2). This range considers the maximal possible angle between two adjoined edges and the fact that the path direction $\phi_{opt}$ is only an average value along the path and is predetermined through the given numbers of possible edges from the differences in slopes according to the chosen mesh structure (see section III.B in [1]). An exact match between the calculated optimal path direction $\phi_{opt}$ and the used edges would be quite a coincidence.

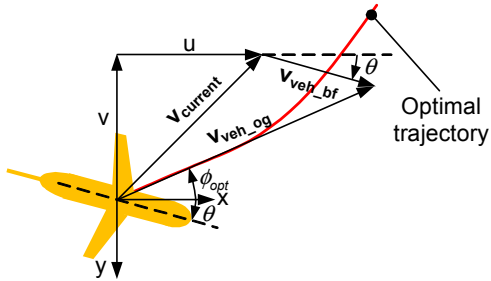

Figure 1. Illustration of the velocities and the angles in glider steering

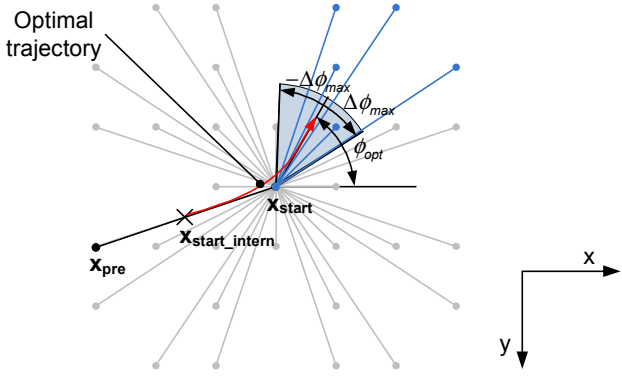

Figure 2. Resultant angle range to define the examined successor vertices using optimal navigation formula from Zermelo

This approach incorporates a pre-selection of promising successor vertices with the goal to decrease the number of cost function calls *wfunc* during the search. Fig. 2 shows the principle idea of the approach using a 3-sector rectangular grid structure which is described in detail in [1] and [2]. By using such a structure, 31 successor vertices are possible from which the approach selects only five. This occurs in the best case (for all examined vertices $v$ is $d[u] < d[v]$) with a resulting reduction of the called *wfunc* to 83 %.

Another possibility in using such an angle range is the symmetrical localization of the angle constraints $\pm\Delta\phi_{max}$ to the direction of the predecessor edge (see Fig. 3). The definition of these constraints should consider the maximal necessary rate of change of the course direction to drive through the current field giving the maximal angle difference between two adjacent edges. The STVE method clearly requires more cost function calls *wfunc* than the approach using the optimal navigation formula. Because here no information exists about a favoured future course direction, the defined course range is so large that all possible changes of course positive or negative are covered.

The determination of the angle constraints for both methods currently use empirical trial search runs with different start positions and a stepwise reduction of the angle range until one of the found paths does not match with the solution using the ITVE algorithm. A future research topic is an automatic detection of these constraints using the information within the grid structure (length of edges, maximum possible angle between two adjoined edges) and the change in gradient of the current.

### D. Calculation of the optimal path direction

The determination of the optimal path direction $\phi_{opt}$ for the previous presented method required a simulation of the optimal trajectory. This simulation is based on a step size control for efficient calculation of numerical solutions of differential equations. This approach was also used in [2] for the calculation of the travel time. The step-size $h$ is not the time as used in numerical solvers but corresponds to the segment lengths of the simulated optimal trajectory and so the optimal trajectory will be shared within many segments. The idea is to simulate the optimal trajectory by starting on the middle position of the previous edge by $\mathbf{x}_{start\_intern}$. The simulation will be stopped if the distance $r_{path}$ between the start position $\mathbf{x}_{start\_intern}$ and the current position $\mathbf{x}_{end\_local}$ is larger than the half length of the previous edge $\|\mathbf{s}_{pre}\|$ and the quarter length of the smallest possible edge $s_{path\_min}$ taken together. This stop criterion was proven to be very successful in the analysis and test of this method. The course, or rather the path direction, of the last simulated segment will be used as the optimal path direction $\phi_{opt}$. (see Fig. 4 for further information). The calculation of the optimal path direction for a segment includes the following steps (see also TABLE IV):

1. Calculation of the travel time $t_{path}$ using the vehicle speed $v_{veh\_bf}$ and the length of the segment $hr_{path}$.
2. Determine the heading rate $\dot{\theta}$ using the optimal navigation formula from Zermelo in equation (2) with the start heading $\theta_{start}$ and the partial derivatives of the ocean current.
3. Calculation of a rough heading $\theta_{end\_rough\_approx}$ used in vector $\mathbf{v}_{veh\_bf}$ using $\theta_{start}$, the travel time $t_{path}$ and heading rate $\dot{\theta}$.
4. Determine the vector of the vehicle speed over ground $\mathbf{v}_{veh\_og}$ using the ocean current vector $\mathbf{c}_{start}$ and the vehicle speed through water vector $\mathbf{v}_{veh\_bf}$ to calculate the end position of the segment $\mathbf{x}_{end\_local}$.
5. Determine the ocean current $\mathbf{c}_{end}$ and its partial derivatives from the endpoint $\mathbf{x}_{end\_local}$ to the time $t_{start\_local} + t_{path}$.
6. Calculation of the average ocean current components along the segment by arithmetic mean of the current components on the start and end position $\mathbf{x}_{start\_local}$ and $\mathbf{x}_{end\_local}$.
7. Improved approximation of the heading $\theta_{end\_improved\_approx}$ on the segment (step 2 and 3) used in the mean partial derivatives of the ocean current.

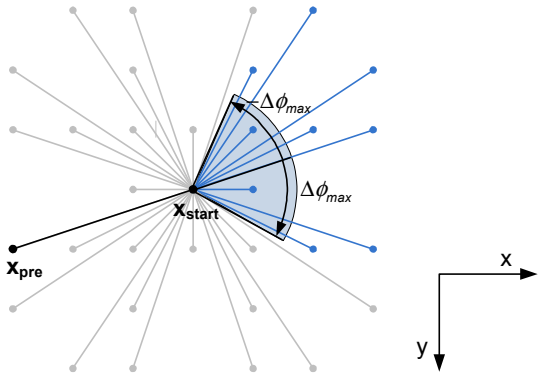

Figure 3. Symmetrical angle range to define the examined successor vertices

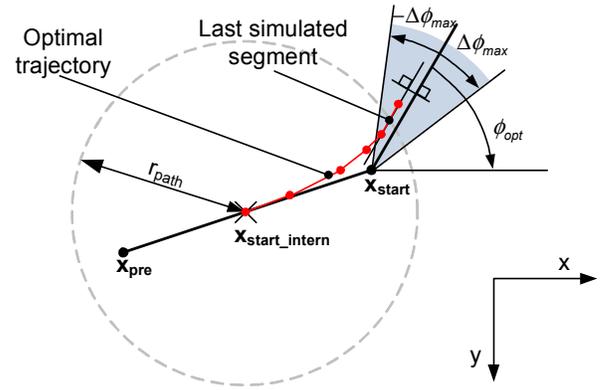

Figure 4. Determine the optimal path direction $\phi_{opt}$

Finally the calculation of the local error $error_{local}$ between the two angles $\theta_{end\_rough\_approx}$ and $\theta_{end\_improved\_approx}$ and the determination of the new step-size $h$ uses the equation for an optimal step-size for a second order method [12]:

$$h = \max\left\{h_{min}, \min\left\{h_{max}, \tau h\sqrt{\frac{\varepsilon}{error_{local}}}\right\}\right\} \quad (3)$$

The parameter $\tau$ is a safety factor ($\tau \in (0, 1]$). Acceptance or rejection of this step is dependent on the local error $error_{local}$ and the calculated step size $h$. TABLE IV includes the details of the algorithm.

TABLE IV
PSEUDO-CODE OF THE ALGORITHM TO CALCULATE THE OPTIMAL COURSE

CALC-OPTCOURSE ($\mathbf{x}_{pre}$, $\mathbf{x}_{start}$, $t_{pre}$, $t_{start}$, $s_{path\_min}$)
*defined parameters:* $v_{veh\_bf}$, $h$, $h_{min}$, $h_{max}$, $\varepsilon$, $\tau$
$t_{start\_local} = 0.5(t_{pre} + t_{start})$
$\mathbf{s}_{pre} = \mathbf{x}_{start} - \mathbf{x}_{pre}$
$\phi_{path} = \mathrm{atan2}(s_{pre}^y, s_{pre}^x)$
$\mathbf{x}_{start\_intern} = \mathbf{x}_{pre} + 0.5\mathbf{s}_{pre}$
$\mathbf{x}_{start\_local} = \mathbf{x}_{start\_intern}$
$r_{travel} = 0$
$r_{path} = 0.25 s_{path\_min} + 0.5\|\mathbf{s}_{pre}\|$
[$\mathbf{c}_{start}$, $\mathbf{u}_{start}$, $\mathbf{v}_{start}$] = GET-CURRENT ($\mathbf{x}_{start\_local}$, $t_{start\_local}$)
$\theta_{start}$ = CALC-COURSE ($\phi_{path}$, $\mathbf{c}_{start}$, $v_{veh\_bf}$)
while ($r_{travel} < r_{path}$)
    $t_{path} = h r_{path}/v_{veh\_bf}$
    $\dot{\theta} = u_{start}^y \cos^2\theta_{start} + (u_{start}^x - v_{start}^y)\cos\theta_{start}\sin\theta_{start} + v_{start}^x \sin^2\theta_{start}$
    $\theta_{end\_rough\_approx} = \theta_{start} + t_{path}\dot{\theta}$
    $\mathbf{v}_{veh\_bf} = v_{veh\_bf}\left[\cos(\theta_{end\_rough\_approx}) \quad \sin(\theta_{end\_rough\_approx})\right]^T$
    $\mathbf{v}_{veh\_og} = \mathbf{c}_{start} + \mathbf{v}_{veh\_bf}$
    $\mathbf{x}_{end\_local} = \mathbf{x}_{start\_local} + t_{path}\mathbf{v}_{veh\_og}$
    [$\mathbf{c}_{end}$, $\mathbf{u}_{end}$, $\mathbf{v}_{end}$] = GET-CURRENT ($\mathbf{x}_{end\_local}$, $t_{start\_local} + t_{path}$)
    $\mathbf{c}_{mean} = 0.5(\mathbf{c}_{start} + \mathbf{c}_{end})$
    $\mathbf{u}_{mean} = 0.5(\mathbf{u}_{start} + \mathbf{u}_{end})$
    $\mathbf{v}_{mean} = 0.5(\mathbf{v}_{start} + \mathbf{v}_{end})$
    $\dot{\theta} = u_{mean}^y \cos^2\theta_{start} + (u_{mean}^x - v_{mean}^y)\cos\theta_{start}\sin\theta_{start} + v_{mean}^x \sin^2\theta_{start}$
    $\theta_{end\_improved\_approx} = \theta_{start} + t_{path}\dot{\theta}$
    $\mathbf{v}_{veh\_bf} = v_{veh\_bf}\left[\cos(\theta_{end\_improved\_approx}) \quad \sin(\theta_{end\_improved\_approx})\right]^T$
    $\mathbf{v}_{veh\_og} = \mathbf{c}_{mean} + \mathbf{v}_{veh\_bf}$
    $\mathbf{x}_{end\_local} = \mathbf{x}_{start\_local} + t_{path}\mathbf{v}_{veh\_og}$
    $error_{local} = |\theta_{end\_rough\_approx} - \theta_{end\_improved\_approx}|$
    $h = \max(h_{min}, \min(h_{max}, \tau h\sqrt{\varepsilon / error_{local}}))$
    if (($error_{local} < \varepsilon$) OR ($h = h_{min}$))
        $\mathbf{c}_{start} = \mathbf{c}_{end}$
        $\mathbf{u}_{start} = \mathbf{u}_{end}$
        $\mathbf{v}_{start} = \mathbf{v}_{end}$
        $\mathbf{x}_{start\_local} = \mathbf{x}_{end\_local}$
        $t_{start\_local} = t_{start\_local} + t_{path}$
        $\theta_{start} = \theta_{end\_improved\_approx}$
        $r_{travel} = \|\mathbf{x}_{end\_local} - \mathbf{x}_{start\_intern}\|$
return $\phi_{opt} = \mathrm{atan2}(v_{veh\_og}^y, v_{veh\_og}^x)$

### E. The use of both methods

The use of both methods together, the A* algorithm (see section B) and the optimal navigation formula from Zermelo (see section C) in the ITVE algorithm combines the two acceleration mechanisms and produces a larger reduction in the computing time than with either method alone. TABLE V shows this algorithm with a few explanations. The modifications to the ITVE algorithm are highlighted. The letters which appear in the explanation column refers to the used method (A*: A* algorithm; Z: Zermelo's formula).

At this point additional acceleration possibilities should be discussed briefly. The first possibly includes the selective reduction of the search area with the goal to decrease the number of examined vertices during the search. To do this a first search run uses a graph with a large grid size and/or a simple grid structure (see [1] and [2]). Around the found path a new geometrical graph will be generated, similar to a pipe. This graph will have a fine grid size and/or a complex grid structure and will be used in a second run to find the optimal path. A modification of the upper approach is the use of a simple cost function in the first search run and the use of an accurate glider-model in the cost function for the second run.

TABLE V
PSEUDO-CODE OF THE ZA*TVE ALGORITHMS

| | Explanations |
|---|---|
| ZA*TVE($G$, $s$, $g$, $t_0$, $\phi_{\Delta max}$) | |
| for each vertex $u \in V$ | |
|   $d[u] \leftarrow \infty$ | |
|   $f[u] \leftarrow \infty$ | A* (initialize heuristic vector) |
|   $\pi[u] \leftarrow \infty$ | |
|   $color[u] \leftarrow$ WHITE | |
| $color[s] \leftarrow$ GRAY | |
| $d[s] \leftarrow t_0$ | |
| $f[s] \leftarrow t_0 + h[s]$ | A* (calculate heuristic for vertex $s$) |
| INSERT($Q$, $s$) | discover vertex $s$ |
| while ($Q \neq \emptyset$) | |
|   $u \leftarrow$ EXTRACT-MIN($Q,f$) | A* examine vertex $u$ |
|   if ($u = g$) | A* (path found) |
|     return ($d$, $\pi$) | A* (program termination) |
|   $color[u] \leftarrow$ BLACK | |
|   if ($u \neq s$) | |
|     $\phi_{opt}$=CALC-OPTDIR($\pi[u],u,d[\pi[u]],d[u]$) | Z (calculate optimal course) |
|   for each $v \in Adj[u]$ | examine edge ($u,v$) |
|     if ($d[u] < d[v]$) | |
|       $\phi_{path}$ = CALC-PATHDIR($u,v$) | Z (calculate edge direction) |
|       if (($u=s$) OR ($\|\phi_{opt}-\phi_{path}\|<\phi_{\Delta max}$)) | Z (select possible successor edges) |
|       $d_v$ = wfunc($u$, $v$, $d[u]$) + $d[u]$ | calculate cost function |
|       if ($d_v < d[v]$) | |
|         $d[v] \leftarrow d_v$ | |
|         $f[v] \leftarrow d_v + h[v]$ | A* (calculate heuristic function) |
|         $\pi[v] \leftarrow u$ | |
|         if ($color[v]$ = GRAY) | |
|           DECREASE-KEY($Q,v,f[v]$) | A* (change heuristic for $v$ in $Q$) |
|         else | |
|           $color[v] \leftarrow$ GRAY | discover ($color[v]$=WHITE) or |
|           INSERT($Q$, $v$) | reopen ($color[v]$=BLACK) vertex $v$ |
| return ($d$, $\pi$) | |

## III. DETECTION OF THE OPTIMAL DEPARTURE TIME

A practice-relevant requirement for optimal path planning for the AUV "SLOCUM Glider" is the determination of the optimal departure time. So is it very difficult to start a glider mission near the coast in the presence of strong tides. Through the low cruising speed (0.2 to 0.4 m s$^{-1}$) and a false chosen start time in combination with a strong flowing tide, it is possible that the glider will make poor forward progress or drift back to the shore. Another scenario is a bad weather situation or a temporary adverse ocean current condition in the region of interest.

### A. Idea

The function to describe the relationship between the travel time $t_{trav}$ and departure time $t_{dep}$ consists of an independent single pair of variants. This means that to determine the travel time for a certain departure time, knowledge of travel times with a lesser departure time is not necessary. Because of this, it is possible to reproduce the principal run of the curve $t_{trav} = f(t_{dep})$ using a smaller number of defined departure times $t_{dep\_i}$, distributed in the time window of interest, to find the corresponding travel times $t_{trav\_i}$. In an additional step the region of the global minimum can be localized, to detect the optimal departure time using a root-finding algorithm. The algorithmic details will be described in the next section.

### B. Algorithm

The detection of the optimal departure time occurs in three steps. Fig. 5 displays an overview of the scheme to determine the optimal departure time. The first step creates supporting points for the curve $t_{trav} = f(t_{dep})$ at intervals of $\Delta t_{dep}$. The choice of the interval width is based on the run of the curve and should reflect the positions of the local minima.

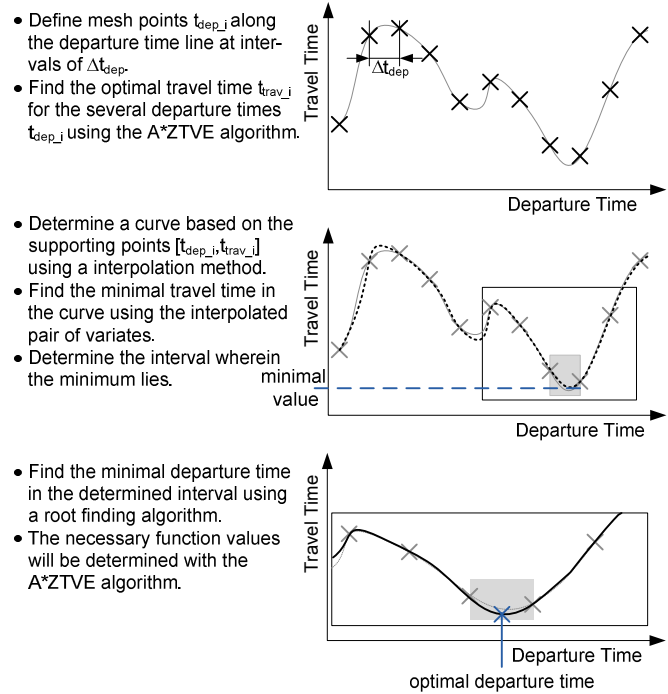

Figure 5. Steps to find the optimal departure time

These supporting points will be provided in a second step to create the approximated run of the curve using an interpolation method. The studies in this research field favour the Akima interpolation [13]. This method provides the best fitting to the real curve and tries to avoid overshoots, which would indicate a nonexistent minimum. The determination of the interval wherein the global minimum of the approximated curve lies is the precondition for the last step.

Here a one-dimensional root-finding algorithm will be used to find the optimal departure time. Thereby a path search using the A*ZTVE algorithm will be running alongside every function call to find the travel time for the given departure time. For root-finding algorithms, root-bracketing algorithms will be used. These algorithms work without derivatives and find the root through iterative decreasing of the interval until a desired tolerance is achieved, wherein the root lies. Golden section search [14], Fibonacci search [15] and Brent's algorithm [16] were tested. Brent's algorithm has the best performance and will be favoured.

### C. Possible Modifications and critical nodes

The above described algorithm calls the search algorithm multiple times, which correlates directly with the processing time. A few possibilities to reduce the processing time will be discussed briefly. Because the localized global minimum in the second step represents only the rough position, the supporting points used in the interpolation do not have to be accurate. This means that in order to detect these points a graph with a larger grid size and/or a simpler grid structure can be used. Another possibility is the use of a simple cost function or a decrease of the angle range $\pm\Delta\phi_{max}$ (see II.C) during the search. These possibilities result in a decrease of the number of examined vertices during the search or lead to a more rapid calculation of the cost function and thus to an acceleration of the computing time.

The multiple calls of the search algorithm can be calculated independently at the same time on separate processor cores of a multi-core computer. The Task library [17], which is a component of the Boost Sandbox [18], provides interesting concepts and functions to solve this challenge, where a thread pool calls the ZA*TVE algorithm for each of the departure times $t_{dep\_i}$ multiple. The analyses of the possibilities for parallelization and the programmable implementation are future work fields.

Use of this approach in a real application requires recognition of the fact that only a limited extent of the forecast window will be available. So, the possible mission window is narrowed down to the period between the considered departure time and the forecast horizon. In the application presented in [3] the forecast window for the ocean currents is ten days. This means that if one starts a mission on the ninth day only a one day mission can be planned. Another aspect is the delayed supply of the data of interest in the case of a later start time. All these points should be considered if one intends to use an optimal departure time in the mission planning. On the other hand, waiting a day for better weather conditions or waiting for a falling tide can also result in a successful mission.

## IV. MAXIMIZING OF THE COLLECTED OCEANOGRAPHIC DATA

In a few applications the main focus is the generation of a path where the glider can collect as much oceanographic data as possible in the region of interest. The time optimality in this case is a secondary consideration. To do this, a new cost function must be defined which rates the position of several path segments to the desired path in the region of interest. This cost function calculates the area between the path segment and the desired path (XTE area). Fig. 6 shows this principal for few path segments. The function *afunc* to calculate this area will be used in the HTTVE (hold track TVE) algorithm instead of the *wfunc* function used in the ITVE algorithm (see TABLE VI).

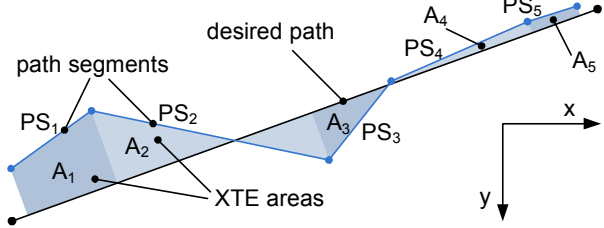

Figure 6. XTE areas for several path segments

TABLE VI
PSEUDO-CODE OF THE HTTVE ALGORITHMS

| ITVE(G, s, $t_0$) | HTTVE(G, s, g, $t_0$, $v_{min}$, $\phi_{\Delta max}$) |
|---|---|
| for each vertex $u \in V$ | for each vertex $u \in V$ |
|  | $a[u] \leftarrow \infty$ |
| $d[u] \leftarrow \infty$ | $d[u] \leftarrow \infty$ |
| $\pi[u] \leftarrow \infty$ | $\pi[u] \leftarrow \infty$ |
| color[u] ← WHITE | color[u] ← WHITE |
| color[s] ← GRAY | color[s] ← GRAY |
|  | $a[s] \leftarrow 0$ |
| $d[s] \leftarrow t_0$ | $d[s] \leftarrow t_0$ |
| INSERT(Q, s) | INSERT(Q, s) |
| while (Q≠Ø) | while (Q≠Ø) |
| $u \leftarrow$ EXTRACT-MIN(Q,d) | $u \leftarrow$ EXTRACT-MIN(Q,a) |
| color[u] ← BLACK | color[u] ← BLACK |
|  | if (u≠s) |
|  | $\phi_{dir}$=CALC-DIR($\pi[u]$,u) |
| for each $v \in Adj[u]$ | for each $v \in Adj[u]$ |
| if (d[u] < d[v]) | if (a[u] < a[v]) |
|  | $\phi_{path}$ = CALC-PATHDIR (u,v) |
|  | if ((u=s) OR ($|\phi_{dir} - \phi_{path}| < \phi_{\Delta max}$)) |
|  | w = wfunc(u, v, d[u]) |
|  | $v_{path}$ = ||pos(v)-pos(u)||/w |
|  | if ($v_{path} > v_{min}$) |
| $d_v$ = wfunc(u, v, d[u]) + d[u] | $a_v$ = afunc(u, v, s, g) + a[u] |
| if ($d_v < d[v]$) | if ($a_v < a[v]$) |
|  | $a[v] \leftarrow a_v$ |
| $d[v] \leftarrow d_v$ | $d[v] \leftarrow w + d[u]$ |
| $\pi[v] \leftarrow u$ | $\pi[v] \leftarrow u$ |
| if (color[v] = GRAY) | if (color[v] = GRAY) |
| DECREASE-KEY(Q,v,$d_v$) | DECREASE-KEY(Q,v,$a_v$) |
| else | else |
| color[v] ← GRAY | color[v] ← GRAY |
| INSERT(Q, v) | INSERT(Q, v) |
| return (d, $\pi$) | return (d, $\pi$) |

## V. RESULTS

### A. The selected test function for a Time-Varying Ocean Flow

The function used to represent a time-varying ocean flow describes a meandering jet in the eastward direction, which is a simple mathematical model of the Gulf Stream [19] and [20]. This function was applied in [1] and [2] to test the TVE algorithm and will be used in the following sections to show the influence of the methods to realize a fast search algorithm and to find an optimal departure time. The stream function is:

$$\phi(x,y) = 1 - \tanh\left(\frac{y - B(t)\cos(k(x-ct))}{\left(1+k^2 B(t)^2 \sin^2(k(x-ct))\right)^{\frac{1}{2}}}\right) \quad (4)$$

which uses a dimensionless function of a time-dependent oscillation of the meander amplitude

$$B(t) = B_0 + \varepsilon \cos(\omega t + \theta) \quad (5)$$

and the parameter set $B_0$ = 1.2, $\varepsilon$ = 0.3, $\omega$ = 0.4, $\theta$ = π/2, $k$ = 0.84 and $c$ = 0.12 to describe the velocity field:

$$u(x,y,t) = -\frac{\partial \phi}{\partial y} \quad v(x,y,t) = \frac{\partial \phi}{\partial x} \quad (6)$$

The dimensionless value for the body-fixed vehicle velocity $v_{veh\_bf}$ is 0.5. The exact solution was founded by solving a boundary value problem (BVP) with a collocation method `bvp6c` [21] in MATLAB. The three ordinary differential equations (ODEs) include the two equations of motion:

$$\begin{aligned} \frac{dx}{dt} &= u + v_{veh\_bf} \cos\theta \\ \frac{dy}{dt} &= v + v_{veh\_bf} \sin\theta \end{aligned} \quad (7)$$

and the optimal navigation formula from Zermelo [10]:

$$\frac{d\theta}{dt} = -u_y \cos^2\theta + (u_x - v_y)\cos\theta \sin\theta + v_x \sin^2\theta . \quad (8)$$

### B. Comparison between the methods to accelerate the TVE algorithm

This section presents the results of the methods to accelerate the TVE algorithm which are described in section II using the time-varying ocean flow test function of the previous section. For the test cases, five different start positions were distributed in the whole area of operation as shown in Fig. 7. All the graph-based methods use the same graph and hence produce identical paths. Fig. 7 shows the five paths found using optimal control and the graph methods. For the graph methods, the rectangular 3-sector grid structure with a grid size of 0.4 was used (see [1] and [2]). Fig. 8 shows the necessary number of cost function calls (CFC) using the several methods for the five start positions. All these results are included in TABLE VII. The examination of the current model calls should reflect their ratio to the cost function calls, which is important in case of computing intensive ocean current calculations [3].

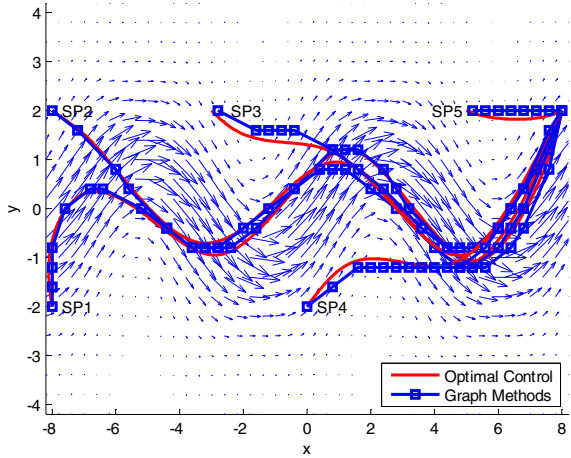

Figure 7. Time optimal paths through a time-varying ocean field using Optimal Control and the Graph Methods for different start positions

The use of the improved TVE algorithm (ITVE) provides a decrease in the number of function calls to about half in comparison to the TVE algorithm which was presented in [1] and [2]. This simple inquiry in the ITVE program is included in the other acceleration methods as well. Using the A* algorithm (A*TVE) (see section II.B), the number of function calls correlates directly with the distance between the start and the goal position. This is reasonable since the algorithm includes only a subset of the vertices in the path search, in fact, only the preferred vertices with a short distance to the goal point. The inclusion of Zermelo's optimal navigation formula in the search algorithm (ZTVE) (see section II.C) results in a decrease of the number of cost function calls to about one quarter of the calls using the ITVE algorithm. With both methods used together (ZA*TVE), the two merged acceleration mechanisms provide a further decrease of the number of cost function and current model calls. The use of the ZA*TVE algorithm allows a decrease of the number of cost function calls (CFC) by about a factor of 12, and, by a factor of 9 for the current model calls (CMC) in comparison to the original TVE algorithm in [1] and [2]. This improvement makes the practical use of the ZA*TVE algorithm possible for the case of (i) the computationally-intensive ocean current calculations as in [3], or, (ii) to determine the optimal departure time.

TABLE VII
RESULTS OF THE DIFFERENT SEARCH METHODS

| Method | SP1 No. of CFC/ No. of CMC | SP2 No. of CFC/ No. of CMC | SP3 No. of CFC/ No. of CMC | SP4 No. of CFC/ No. of CMC | SP5 No. of CFC/ No. of CMC |
|---|---|---|---|---|---|
| TVE | 24008/ 160857 | 24122/ 163236 | 24262/ 162596 | 24262/ 165379 | 24013/ 162584 |
| ITVE | 12124/ 80838 | 12126/ 80726 | 12147/ 81635 | 12147/ 83886 | 12112/ 83757 |
| A*TVE | 7629/ 46916 | 6718/ 38564 | 4042/ 24668 | 2860/ 16673 | 638/ 5559 |
| ZTVE | 3076/ 27734 | 2953/ 26842 | 2763/ 25211 | 2817/ 25449 | 2824/ 25295 |
| ZA*TVE | 1883/ 17502 | 1678/ 15630 | 934/ 8467 | 627/ 5097 | 141/ 1409 |

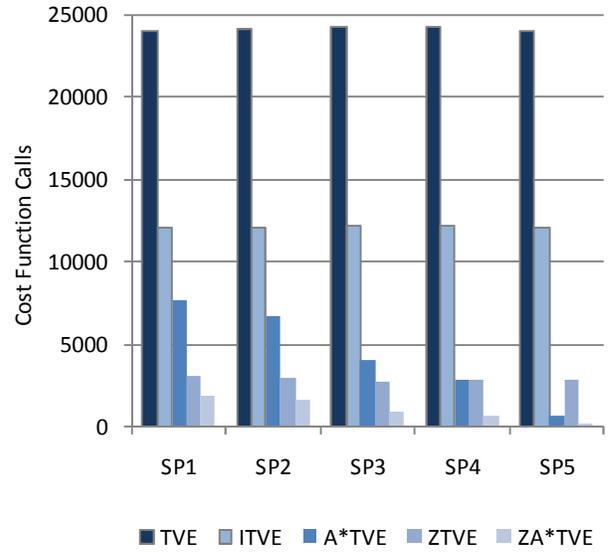

Figure 8. Cost function calls for the various methods with different start positions

### C. Necessary Function Calls to Determine the Optimal Departure time

An analysis of the necessary function calls for each of the program steps to determine the optimal departure time (see III.B) will be presented in this section. Thus several different acceleration methods and root-finding algorithms will be considered. For the test cases, start position SP1 in Fig. 7, a graph with a rectangular 3-sector grid structure, and a grid size of 0.4 will be used (see [1]). The applicable angle range in the ZA*TVE algorithm is 27.5°. The results are included in TABLE VIII. Fig. 9 shows the real run of the curve as well as the interpolated curve based on the supporting points. The real curve is determined with an interval $\Delta t_{dep} = 0.01$. The chosen interval to create the approximated run of the curve is $\Delta t_{dep} = 4$. The first acceleration possibility which was tested includes the decrease of the angle range from 27.5° to 17.5° to restrict possible successor edges in the ZA*TVE algorithm during the search. This modification provides a suboptimal solution because few necessary optimal edges will be disregarded.

TABLE VIII
NECESSARY FUNCTION CALLS USING DIFFERENT ROOT-FINDING ALGORITHM AND ACCELERATION METHODS

| Search Method | Optimal Departure Time | Optimal Travel Time | No. of Search Calls | No. of CFC | No. of CMC |
|---|---|---|---|---|---|
| Approximate minimum | | | | | |
| Without acceleration | 42.69 | 14.370 | 22 | 46610 | 431284 |
| Decrease angle range | 43.81 | 14.425 | 22 | 26664 | 266088 |
| Simpler grid structure | 42.84 | 14.583 | 22 | 23822 | 219805 |
| Exact minimum | | | | | |
| Golden section search | 42.591 | 14.291 | 15 | 19752 | 184686 |
| Fibonacci search | 42.585 | 14.291 | 13 | 17130 | 160185 |
| Brent's algorithm | 42.591 | 14.291 | 9 | 11872 | 110993 |
| Best combination | | | | | |
| Without acceleration | 42.591 | 14.291 | 31 | 58482 | 542277 |
| With acceleration | 42.591 | 14.291 | 31 | 35694 | 330798 |

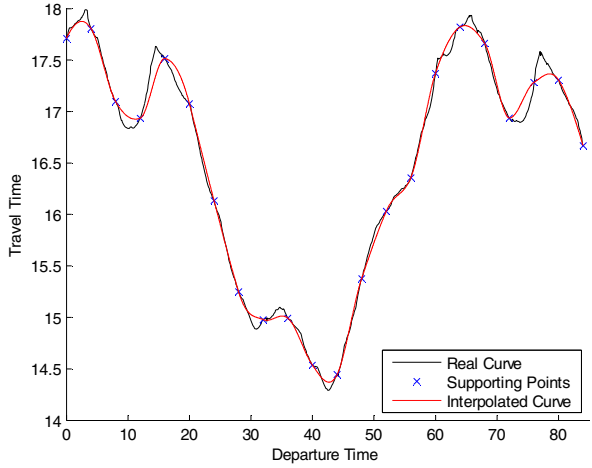

Figure 9. Real and interpolated curve to describe the functional relationship between the travel time and the departure time using the real

The search algorithm builds the path with neighbouring edges to reproduce the optimal edges and so creates a suboptimal path. The use of a rectangular 2-sector grid structure to create the approximated run of the curve provides the best acceleration possibility. This method provides the largest saving in the number of the cost function calls.

To find the exact minimum, three one-dimensional root–finding algorithms were tested. These are the Golden section search, Fibonacci search and Brent's algorithm. The termination tolerance is 0.01. The smallest number of search calls to find the minimum requires Brent's algorithm. The best combination of the favoured method to find the approximated minimum (simpler grid structure) and the exact minimum (Brent's algorithm) needs 19 times more function calls than the ZA*TVE algorithm and only two times more than the original TVE algorithm in [1].

## VI. BENCHMARKS

This section presents a few benchmark tests of the various algorithms using the mathematical model in section V.A as well as real netCDF files for a 10-day forecast which is described in [3]. The compiled C++ programs were tested with an Intel Core 2 Extreme CPU Q9300 @ 2.53 GHz. The full performance of the multi-core processor cannot be used, because the programs are not presently coded in parallel.

TABLE IX includes the computing time for the various search algorithms using the Gulf Stream model with different grid sizes. The acceleration factor of the computing time for the various algorithms doesn't correspond directly with the number of calls to the cost function. The reasons are the additional programming steps of the A*TVE or ZTVE algorithm in comparison to the ITVE or STVE algorithm, as well as the simple cost function calculation in comparison to the additional programming steps of the search algorithms. These are the creation of the data structures and the storage, sort and search of elements in the structures. The use of the ZA*TVE algorithm allows a decrease of the calculating time by about a factor 2.7 to 3.7 in comparison to the original TVE algorithm.

TABLE IX
RESULTS OF THE DIFFERENT SEARCH METHODS USING

| Method | Rectangle 3-sector Grid size | No. of CFC/ | No. of CMC | Computing Time in ms | | |
|---|---|---|---|---|---|---|
| | | | | Generate Graph | Path Search | Total Time |
| TVE | 0.4 | 24296 | 160857 | 6.2 | 68.4 | 64.6 |
| | 0.05 | 1627856 | 5821769 | 441.6 | 2062.9 | 2504.5 |
| ITVE | 0.4 | 12124 | 80838 | 6.0 | 30.2 | 36.2 |
| | 0.05 | 813928 | 2893716 | 398.2 | 1055.2 | 1453.4 |
| A*TVE | 0.4 | 7629 | 46916 | 6.0 | 25.3 | 31.3 |
| | 0.05 | 471858 | 1586213 | 393.8 | 1001.1 | 1394.9 |
| STVE | 0.4 | 4354 | 33880 | 5.8 | 15.8 | 21.6 |
| | 0.05 | 358744 | 1284032 | 402.8 | 686.3 | 1089.1 |
| ZTVE | 0.4 | 3076 | 27734 | 6.0 | 14.8 | 20.8 |
| | 0.05 | 251190 | 1047517 | 389.6 | 646.6 | 1036.2 |
| SA*TVE | 0.4 | 2729 | 21645 | 6.1 | 13.9 | 20.0 |
| | 0.05 | 202864 | 718042 | 403.6 | 563.6 | 967.2 |
| ZA*TVE | 0.4 | 1883 | 17502 | 5.9 | 11.4 | 17.3 |
| | 0.05 | 140338 | 578104 | 400.5 | 536.9 | 937.4 |

The results of the search algorithms to create real mission plans along the Newfoundland Shelf using an ocean current model to extract the ocean current information from netCDF data files (see [3]) are included in TABLE X. Because the partial derivatives of the ocean current components will not be provided from the ocean current model, only the symmetrical angle range to define the examined successor vertices will be used (STVE algorithm). The acceleration factor between the TVE and the SA*TVE algorithm is, in this application, 5 to 7.6.

TABLE XI and Fig. 10 show the results of the HTTVE algorithm in section IV using various minimal vehicle speeds $v_{min}$ in the previous practical test application. It is clearly shown that the resulting paths are matched well with the desired path to collect as much oceanographic data as possible, which is a straight line from the start to the goal position. Only in the case of an adverse current, will the path depart significantly from that straight line (see M31 and M32).

TABLE X
RESULTS OF THE DIFFERENT SEARCH METHODS

| Mission | | M31 | M32 | M33 | M34 |
|---|---|---|---|---|---|
| Travel Time D:h:min:s | | 08:06:15:55 | 07:06:12:24 | 08:05:24:48 | 07:14:30:37 |
| Path Length km | | 230.12 | 220.66 | 213.95 | 218.55 |
| No. of Vertices | | 9462 | 9462 | 5125 | 9462 |
| No of Edges | | 292238 | 292238 | 155128 | 292238 |
| Computing Time in s / No. of Cost Function Calls 3D / No. of Cost Function Calls / No of Current Function Calls | TVE | 82.6 / 292238 / 2523945 / 19408534 | 92.8 / 292238 / 2526954 / 19433223 | 48.7 / 155128 / 1348077 / 10324120 | 96.8 / 292238 / 2690365 / 20110589 |
| | ITVE | 42.9 / 146116 / 1266566 / 9744202 | 47.7 / 146119 / 1282297 / 9883492 | 25.3 / 75562 / 684707 / 5272871 | 48.5 / 146119 / 1353200 / 10126879 |
| | A*TVE | 31.7 / 95759 / 844992 / 6457420 | 26.9 / 81235 / 720295 / 5482202 | 21.7 / 62280 / 574127 / 4411799 | 35.6 / 106008 / 988037 / 7350671 |
| | STVE | 18.9 / 60389 / 555183 / 4306693 | 21.6 / 61198 / 568687 / 4419721 | 11.4 / 31866 / 299167 / 2318107 | 21.5 / 62269 / 595393 / 4457650 |
| | SA*TVE | 14.2 / 39594 / 368992 / 2842646 | 12.1 / 33759 / 316216 / 2427513 | 9.7 / 25894 / 245884 / 1899355 | 15.8 / 44853 / 431018 / 3204780 |

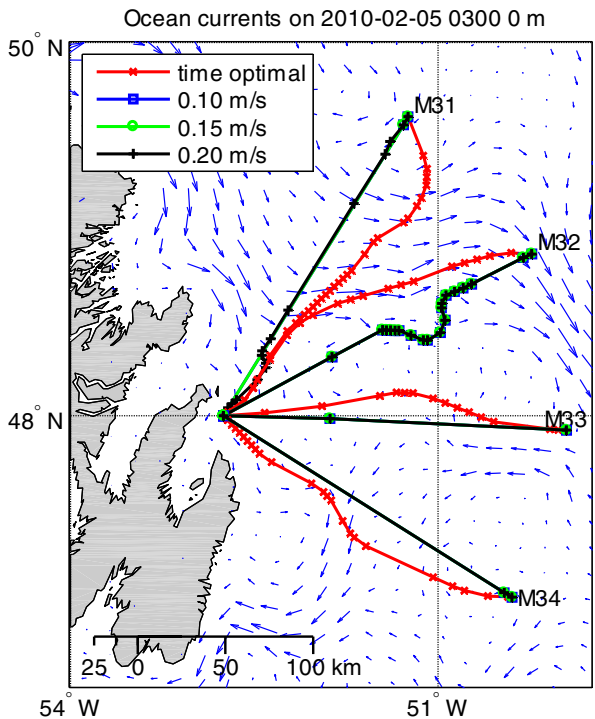

Figure 10. Time optimal path for different missions

TABLE XI
RESULTS FOR VARIOUS MINIMAL VEHICLE SPEEDS

| Mission | Travel Time Path d:h:min:s | Travel Time Straight Line d:h:min:s | Path Length Path km | Path Length Straight Line km |
|---|---|---|---|---|
| M31 time optimal | 08:06:16:55 | 09:18:22:03 | 230.12 | 210.55 |
| M31 $v_{min}$=0.10 m/s | 09:18:19:34 | 09:18:22:03 | 210.56 | 210.55 |
| M31 $v_{min}$=0.15 m/s | 09:18:19:34 | 09:18:22:03 | 210.56 | 210.55 |
| M31 $v_{min}$=0.20 m/s | 09:20:11:07 | 09:18:22:03 | 214.89 | 210.55 |
| M32 time optimal | 07:06:12:24 | NaN | 220.66 | 210.55 |
| M32 $v_{min}$=0.10 m/s | 08:23:11:29 | NaN | 228.09 | 210.55 |
| M32 $v_{min}$=0.15 m/s | 08:23:11:29 | NaN | 228.09 | 210.55 |
| M32 $v_{min}$=0.20 m/s | 08:23:11:29 | NaN | 228.09 | 210.55 |
| M33 time optimal | 08:05:24:48 | 08:12:54:13 | 213.95 | 210.00 |
| M33 $v_{min}$=0.10 m/s | 08:12:54:13 | 08:12:54:13 | 210.00 | 210.00 |
| M33 $v_{min}$=0.15 m/s | 08:12:54:13 | 08:12:54:13 | 210.00 | 210.00 |
| M33 $v_{min}$=0.20 m/s | 08:12:54:13 | 08:12:54:13 | 210.00 | 210.00 |
| M34 time optimal | 07:14:30:37 | 08:00:03:31 | 218.55 | 210.55 |
| M34 $v_{min}$=0.10 m/s | 08:00:00:00 | 08:00:03:31 | 210.56 | 210.55 |
| M34 $v_{min}$=0.15 m/s | 08:00:00:00 | 08:00:03:31 | 210.56 | 210.55 |
| M34 $v_{min}$=0.20 m/s | 08:00:00:00 | 08:00:03:31 | 210.56 | 210.55 |

## VII. CONCLUSIONS

In this paper a few important practice-oriented requirements for optimal path planning for the AUV "SLOCUM Glider" as well as the solutions using fast graph based algorithms are presented. The first part of this paper describes a few methods to accelerate the processing time of the TVE algorithm, which was described in [1] and [2]. The realization of a fast calculation of the algorithm is a precondition for the detection of the optimal departure time, which is presented in the middle part of this paper. The description of an algorithm to generate a path where the glider can collect as much oceanographic data as possible in the region of interest is an additional point of discussion. The last part of this paper shows some test scenarios, which demonstrate the performance of the algorithms in a time-varying ocean field.

ACKNOWLEDGMENT

This work is financed by the German Research Foundation (DFG) within the scope of a two-year research fellowship. I would like to thank the National Research Council Canada Institute for Ocean Technology and in particular Dr. Christopher D. Williams for support during this project. I gratefully acknowledge Dr. Michaël Soulignac from the ISEN engineering school in Lille for the helpful discussions.